\pdfoutput=1
\documentclass[11pt]{article}

\usepackage{ACL2023}

\usepackage{times}
\usepackage{latexsym}
\usepackage{graphicx}

\usepackage[T1]{fontenc}

\usepackage[utf8]{inputenc}
\usepackage{microtype}
\usepackage{inconsolata}

\title{From Superficial Patterns to Semantic Understanding: Fine-Tuning Language Models on Contrast Sets}

\author{Daniel Petrov \\
  University of Texas at Austin \\
  \texttt{dpetrov@utexas.edu} 
}

\begin{document}
\maketitle
\begin{abstract}
Large-scale pre-trained language models have demonstrated high performance on standard datasets for natural language inference (NLI) tasks. Unfortunately, these evaluations can be misleading, as although the models can perform well on in-distribution data, they perform poorly on out-of-distribution test sets, such as contrast sets. Contrast sets consist of perturbed instances of data that have very minor, but meaningful, changes to the input that alter the gold label, revealing how models can learn superficial patterns in the training data rather than learning more sophisticated language nuances. As an example, the ELECTRA-small language model achieves nearly 90\% accuracy on an SNLI dataset but drops to 75\% when tested on an out-of-distribution contrast set. The research carried out in this study explores how the robustness of a language model can be improved by exposing it to small amounts of more complex contrast sets during training to help it better learn language patterns. With this approach, the model recovers performance and achieves nearly 90\% accuracy on contrast sets, highlighting the importance of diverse and challenging training data.
\end{abstract}

\section{Introduction}

This paper focuses on investigating and reducing biases in language models in Natural Language Inference (NLI) tasks. In NLI, the target is to determine the relationship between a premise and a hypothesis and classify whether the premise entails, contradicts, or is neutral compared to the hypothesis. Being able to correctly understand such relationships is fundamental in understanding and building on natural language. Modern state-of-the-art deep learning models, particularly transformer-based architectures, have scored high on such tasks, often even higher than humans.

\begin{figure}[ht]
    \centering
    \includegraphics[width=0.9\linewidth]{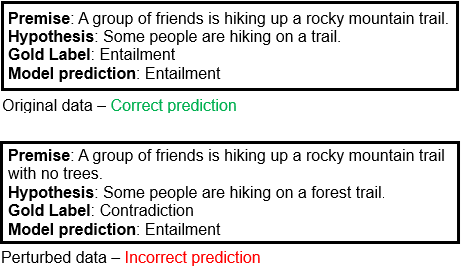} 
    \caption{Comparison between the original data and the perturbed data, showing how the model prediction accuracy decreases from simple changes to the premise and hypothesis.}
    \label{fig:contrast-ex}
\end{figure}

However, recent research has raised the question – \textit{are these models being evaluated properly?} Standard test sets primarily evaluate in-distribution generalization, and if these datasets have systematic gaps (e.g. annotation artifacts), then the evaluations are misleading, as the model may simply be learning decision rules that may perform well on the test set and are not really creating an understanding of language nuances \citep {gardner2020evaluatingmodelslocaldecision}. In other words, the model may be learning spurious correlations or superficial patterns in the training data that can be carried over into the test set, ultimately inflating the accuracy values. Such reliance on shallow heuristics limits the broader applicability of these models, especially in real-world out-of-distribution scenarios.

A more rigorous evaluation approach is to use contrast sets – a collection of minimally perturbed data examples built from the training set. The perturbations are created in a way that alters the correct label of the original example without introducing significant syntactical changes to the input data. The goal of such a set is to evaluate the robustness and generalization capability of the model and test if it can handle slight variations in the input and still produce correct results. As expected, when testing the trained model on a contrast set, the accuracy drops to 74.9\%, showing that the SNLI test set alone is not sufficient to measure performance.

To enhance the model’s generalization capabilities, subsets of the contrast dataset were used to fine-tune the pre-trained model which was then evaluated on the remaining portion of the contrast set. By exposing the model to more complex examples, the model becomes more flexible. The results are promising, increasing accuracy to 90.7\%.

This analysis highlights how critical the training data diversity is in developing models that can not only excel on standard benchmarks but also demonstrate a robust understanding of nuanced language contexts. By using contrast sets, we highlight a path toward building models that are both high-performing and resilient when faced with complexity and variability.

\section{Model \& Dataset}
\subsection{Model}
This paper trains the ELECTRA-small model, a popular model with 14M parameters that is widely used to pre-train transform networks designed specifically for natural language processing tasks \citep{clark2020electrapretrainingtextencoders}. It is computationally efficient and also much smaller than its base and large counterparts (110M and 335M parameters, respectively). The model was trained on a custom machine with an NVIDIA GeForce RTX 4070 GPU.

\subsection{Training and Testing Data}
The popular Stanford Natural Language Inference (SNLI) corpus was used for training and initial testing \citep{bowman2015largeannotatedcorpuslearning}, which is a large collection of 570k human-written English sentence pairs that were manually labeled as entailment, contradiction, or neutral. 

Each example in SNLI comes with a premise sentence, a hypothesis sentence, and a gold label (entailment, contradiction, or neutral). Examples of inputs can be seen in Table~\ref{table:nli}. The goal of the model is to determine whether the premise entails the hypothesis or contradicts it. In some cases, it may also be neutral. The data was split with 550k examples being used for training and 10k used for testing. 

\begin{table}[h]
\centering
\resizebox{\columnwidth}{!}{%
\begin{tabular}{|p{4cm}|p{4cm}|c|}
\hline
\textbf{Premise} & \textbf{Hypothesis} & \textbf{Label} \\ \hline
A young student doing her math homework. & A student does schoolwork. & 0 (Entailment) \\ \hline
Children smiling and waving at a camera. & They are smiling at their parents. & 1 (Neutral) \\ \hline
A football player holding a football. & The football player has empty hands. & 2 (Contradiction) \\ \hline
\end{tabular}%
}
\caption{Examples of Natural Language Inference (NLI) data with premise, hypothesis, and label.}
\label{table:nli}
\end{table}

\subsection{Contrast Set}
The contrast set was automatically generated using Linguistically-Informed Transformations (LIT) \citep{li2020linguisticallyinformedtransformationslitmethod} where labels $y \neq y'$. A total of 14,363 contrastive examples were generated, where 20\% of those examples were used to fine-tune the existing model and the remaining 80\% were used for validation testing. The contrast set contains perturbed examples from the original SNLI data that change the hypothesis in a way that changes the resulting entailment label. Examples can be including antonyms, adding negations, synonyms, semantic changes, etc., and can be seen in Figure~\ref{fig:contrast-ex}.

\section{Method}
The ELECTRA-small model was trained for NLI tasks on the SNLI dataset. During validation of the model on the SNLI test set, it scored a respectable accuracy of 89.8\%. However, after performing the same evaluation on the contrast set, the accuracy dropped significantly by 14.9 points to a total of 74.9\%.

It is important to understand the performance discrepancy and why the model misclassifies examples much more often on the contrast set vs. on the original SNLI data, particularly finding which language structures the model finds difficult. A comprehensive error analysis was performed where 20 incorrectly classified examples were studied and grouped into different categories. The following error categories were used, taken from prior research \citep{naik2018stresstestevaluationnatural}. 

\begin{enumerate}
    \item \textbf{Word Overlap} – High lexical overlap between the premise and hypothesis could trick the model into predicting entailment even though they are unrelated or contradictory.
    \item \textbf{Negation} – Words such as “no” or “not” in the hypothesis could cause the model to incorrectly predict contradiction in cases where it is actually neutral or entailed.
    \item \textbf{Length Mismatch} – The premise contains more words than the hypothesis which could confuse the model and create distractions, leading to incorrect classification.
    \item \textbf{Ambiguity} – Instances where the correct answer is unclear, even to the annotators.
    \item \textbf{Unknown} – It is unclear what caused the model to misclassify.
\end{enumerate}

\begin{table*}[t]
\centering
\resizebox{\textwidth}{!}{%
\begin{tabular}{|p{3cm}|p{5cm}|p{5cm}|c|c|}
\hline
\textbf{Error} & \textbf{Premise} & \textbf{Hypothesis} & \textbf{Gold Label} & \textbf{Predicted Label} \\ \hline
Word Overlap & People standing at a beach with cameras. & People standing at a beach filmed by cameras. & 1 (Neutral) & 0 (Entailment) \\ \hline
Negation & The company no longer manufactures the smartphone model due to declining demand. & The company does not manufacture the smartphone model. & 0 (Entailment) & 2 (Contradiction) \\ \hline
Length Mismatch & A Caucasian man wearing red plaid bottoms and a black vest with large buttons with a tattoo on one arm has an upside-down unicycle balancing on his mouth. & A man is riding a unicycle. & 2 (Contradiction) & 0 (Entailment) \\ \hline
Ambiguity & The man saw the woman with binoculars. & The woman had binoculars. & 1 (Neutral) & 0 (Entailment) \\ \hline
Unknown & The artist created a masterpiece. & The masterpiece was created by the artist. & 0 (Entailment) & 1 (Neutral) \\ \hline
\end{tabular}%
}
\caption{Examples of common error types, including their premise, hypothesis, gold label, and predicted label.}
\label{table:errors2}
\end{table*}

Examples of the error categories are shown in Table~\ref{table:errors2} and of those examples evaluated, the aggregate statistics can be seen in Table~\ref{table:error-comparison}. Word overlap is the highest cause for error, further suggesting that the model is learning patterns in the data rather than true language understanding. Instead of reasoning about the actual semantic relationship between the premise and the hypothesis, the model relies instead on the degree of word overlap for its decision, which is a superficial heuristic that needs to be amended.

\section{Model Improvement}

After training the model on the original SNLI dataset and evaluating on both the test contrast sets, the objective was to enhance the model's robustness and generalization capabilities in order to achieve better results on the contrast set while also getting better or similar results on the original SNLI data. These improvements are essential for addressing limitations in language models that rely on patterns in training data as opposed to developing a deeper understanding of language.

\begin{figure*}[t]
    \centering
    \includegraphics[width=0.75\textwidth]{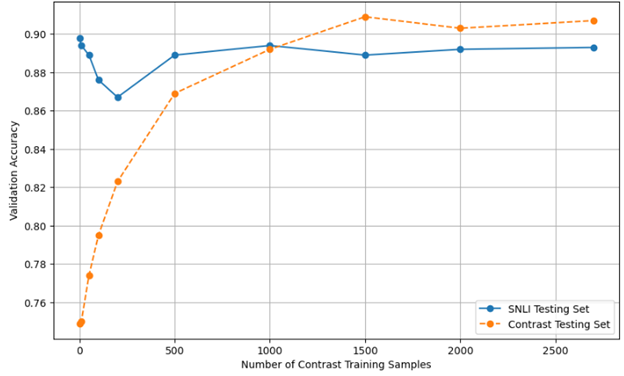} 
    \caption{Performance improvement of the model on contrast sets with varying training sample sizes.}
    \label{fig:improvement}
\end{figure*}

The proposed method is to go through a fine-tuning process where the pre-trained model continues to be trained using a small set of examples taken from the contrast set. The aim was to expose the model to more complex and nuanced language scenarios, allowing it to better learn semantic relationships and context-dependent reasoning. During training, the accuracy on the contrast set was calculated at regular intervals, as shown in Figure~\ref{fig:improvement}. We can see an exponential boost in performance when increasing the number of contrast examples used in training. At approximately 1500 examples used, the accuracy level begins to level off, showing that even using as little as 10\% of the contrast set during training can lead to tremendous improvements in performance. 

When evaluating the fine-tuned model, the number of errors related to superficial patterns in the data (word overlap, negation, etc.) reduces significantly, as shown in Table~\ref{table:error-comparison}. This underscores the initial observation that the model was primarily relying on patterns in the data rather than its genuine understanding of the language. The errors the model faced after tuning are more related to ambiguity, which is more of an issue with the annotated contrast set rather than with the model itself. 

\begin{table}[ht]
\centering
\resizebox{\linewidth}{!}{%
\begin{tabular}{|l|c|c|}
\hline
\textbf{Error Type} & \textbf{Frequency, Before} & \textbf{Frequency, After} \\ \hline
Word Overlap        & 35\%                                     & 10\%                                \\ \hline
Negation            & 20\%                                     & 10\%                                \\ \hline
Length Mismatch     & 15\%                                     & 20\%                                \\ \hline
Ambiguity           & 10\%                                     & 30\%                                \\ \hline
Unknown             & 25\%                                     & 30\%                                \\ \hline
\end{tabular}%
}
\caption{Comparison of error rates for different error types before and after fine-tuning, manually reviewed on 20 incorrectly predicted examples.}
\label{table:error-comparison}
\end{table}

The results of this method were very promising, as the model was able to increase its accuracy on the contrast set by a total of 16.8 points, reaching upward of 90.7\% accuracy. This large boost in performance demonstrates the effectiveness of incorporating even a small subset of more complex, out-of-distribution data to the model. In addition, the increased accuracy on the contrast set did not come at the expense of the accuracy on the original SNLI dataset. The findings are summarized in Table~\ref{table:errors}.

\begin{table}[ht]
\centering
\begin{tabular}{|l|l|c|}
\hline
\textbf{Model} & \textbf{Test Set} & \textbf{Accuracy} \\ \hline
Without fine-tuning & SNLI           & 89.8             \\ 
                    & Contrast Set   & 74.9             \\ \hline
With fine-tuning    & SNLI           & 89.3             \\ 
                    & Contrast Set   & 90.7             \\ \hline
\end{tabular}
\caption{Accuracy of the model on SNLI and Contrast Sets before and after fine-tuning.}
\label{table:errors}
\end{table}

\section{Conclusion}
This paper gives an analysis on how to evaluate language models and highlights the importance of addressing biases and why contrast sets are difficult for models to handle. The ELECTRA-small model initially showed strong performance on in-distribution data, but then significantly dropped on the contrast set, highlighting its susceptibility to spurious correlations in the training data. 

Fine-tuning the pre-trained model with a subset of contrast set examples showed significant gains in accuracy on more complex test sets.  The improvement suggests that training sets with diverse examples can enhance a model’s ability to handle more nuanced language examples, ultimately reducing its bias and increasing its generalization abilities.

\section*{Limitations}
This method assumes the contrast set contains a variety linguistic examples with a balanced label distribution. A skewed dataset could result in the model simply predicting the majority label, regardless of context. Nevertheless, this paper highlights the proof of concept of training on complex data examples to ensure models are robust and able to generalize well.

\bibliography{custom}

\begin{thebibliography}{5}
\expandafter\ifx\csname natexlab\endcsname\relax\def\natexlab#1{#1}\fi

\bibitem[{Bowman et~al.(2015)Bowman, Angeli, Potts, and Manning}]{bowman2015largeannotatedcorpuslearning}
Samuel~R. Bowman, Gabor Angeli, Christopher Potts, and Christopher~D. Manning. 2015.
\newblock \href {http://arxiv.org/abs/1508.05326} {A large annotated corpus for learning natural language inference}.

\bibitem[{Clark et~al.(2020)Clark, Luong, Le, and Manning}]{clark2020electrapretrainingtextencoders}
Kevin Clark, Minh-Thang Luong, Quoc~V. Le, and Christopher~D. Manning. 2020.
\newblock \href {http://arxiv.org/abs/2003.10555} {Electra: Pre-training text encoders as discriminators rather than generators}.

\bibitem[{Gardner et~al.(2020)Gardner, Artzi, Basmova, Berant, Bogin, Chen, Dasigi, Dua, Elazar, Gottumukkala, Gupta, Hajishirzi, Ilharco, Khashabi, Lin, Liu, Liu, Mulcaire, Ning, Singh, Smith, Subramanian, Tsarfaty, Wallace, Zhang, and Zhou}]{gardner2020evaluatingmodelslocaldecision}
Matt Gardner, Yoav Artzi, Victoria Basmova, Jonathan Berant, Ben Bogin, Sihao Chen, Pradeep Dasigi, Dheeru Dua, Yanai Elazar, Ananth Gottumukkala, Nitish Gupta, Hanna Hajishirzi, Gabriel Ilharco, Daniel Khashabi, Kevin Lin, Jiangming Liu, Nelson~F. Liu, Phoebe Mulcaire, Qiang Ning, Sameer Singh, Noah~A. Smith, Sanjay Subramanian, Reut Tsarfaty, Eric Wallace, Ally Zhang, and Ben Zhou. 2020.
\newblock \href {http://arxiv.org/abs/2004.02709} {Evaluating models' local decision boundaries via contrast sets}.

\bibitem[{Li et~al.(2020)Li, Shengshuo, Liu, Wu, Zhou, and Steinert-Threlkeld}]{li2020linguisticallyinformedtransformationslitmethod}
Chuanrong Li, Lin Shengshuo, Leo~Z. Liu, Xinyi Wu, Xuhui Zhou, and Shane Steinert-Threlkeld. 2020.
\newblock \href {http://arxiv.org/abs/2010.08580} {Linguistically-informed transformations (lit): A method for automatically generating contrast sets}.

\bibitem[{Naik et~al.(2018)Naik, Ravichander, Sadeh, Rose, and Neubig}]{naik2018stresstestevaluationnatural}
Aakanksha Naik, Abhilasha Ravichander, Norman Sadeh, Carolyn Rose, and Graham Neubig. 2018.
\newblock \href {http://arxiv.org/abs/1806.00692} {Stress test evaluation for natural language inference}.

\end{thebibliography}
\bibliographystyle{acl_natbib}

\end{document}